\documentclass{article}

\usepackage{arxiv}

\usepackage[utf8]{inputenc} 
\usepackage[T1]{fontenc}    
\usepackage[affil-it]{authblk}
\usepackage{hyperref}       
\usepackage{url}            
\usepackage{booktabs}       
\usepackage{amsfonts}       
\usepackage{nicefrac}       
\usepackage{microtype}      
\usepackage{lipsum}
\usepackage{graphicx}
\graphicspath{ {./images/} }
\usepackage{multirow}%
\usepackage{amsmath,amssymb,amsfonts}%
\usepackage{amsthm}%
\usepackage{mathrsfs}%
\usepackage[title]{appendix}%
\usepackage{xcolor}%
\usepackage{textcomp}%
\usepackage{manyfoot}%
\usepackage{booktabs}%
\usepackage{algorithm}%
\usepackage{algorithmicx}%
\usepackage{algpseudocode}%
\usepackage{listings}%
\usepackage{tabularx}%
\usepackage{siunitx}%
\usepackage{csquotes}%
\usepackage{float}%
\usepackage{babel}%
\usepackage{kotex}%
\usepackage{natbib}%

\usepackage{polyglossia} 
\setmainlanguage{english} 
\setotherlanguage{arabic}
\newfontfamily\arabicfont[Script=Arabic, Extension=.ttf]{NotoNaskhArabic-Regular}

\title{Cultural Perspectives and Expectations for Generative AI: A Global Survey Approach}

\author[1]{Erin van Liemt\thanks{Corresponding author: \href{mailto:evanliemt@google.com}{evanliemt@google.com}}}
\author[1]{Renee Shelby}
\author[1]{Andrew Smart}
\author[2]{Sinchana Kumbale\protect\\}
\author[3]{Richard Zhang}
\author[2]{Neha Dixit}
\author[1]{Qazi Mamunur Rashid}
\author[1]{Jamila Smith Loud}

\affil[1]{Google Research, USA}
\affil[2]{Google, India}
\affil[3]{Google DeepMind, USA}

\begin{document}
\maketitle
\begin{abstract}
There is a lack of empirical evidence about global attitudes around whether and how GenAI should represent cultures. This paper assesses understandings and beliefs about culture as it relates to GenAI from a large-scale global survey. We gathered data about what culture means to different groups, and about how GenAI should approach the representation of cultural artifacts, concepts, or values.  
We distill working definitions of culture directly from these communities to build an understanding of its conceptual complexities and how they relate to representations in Generative AI. 
We survey from across parts of Europe, North and South America, Asia, and Africa. We conclude with a set of recommendations for Culture and GenAI development. These include participatory approaches, prioritizing specific cultural dimensions beyond geography, such as religion and tradition, and a sensitivity framework for addressing cultural ``redlines''.
\end{abstract}


\section{Introduction}
Generative AI (GenAI) applications, such as text-to-image generation, are becoming global technologies integrated into domains such as advertising, content creation, and many others. However, scholars have underscored that significant gaps remain in the cultural representation and understanding capabilities of these systems  (see: \citep{prabhakaran2022cultural, pawar2025culture, johnson2022ghost, khan2025randomness, kirk2024prism, vargaslarge, naik2023social, 11071263}). 
There's a body of research that argues current GenAI development often mirrors the extractive dynamics of historical colonialism \citep{alenichev2025we},
where models are built by a minority of researchers in the Global North and deployed globally without adequate cultural competency \citep{qadri2025risks, adilazuarda2024towards}. 
This results in biases and knowledge gaps that lead to the stereotyping or erasure of non-Western cultures, largely because models are trained on majority-English internet datasets that lack cultural nuance. 


As GenAI adoption increases globally, the distortion or misrepresentation of sensitive cultural symbols and artifacts remains a persistent challenge \citep{qadri2025risks, adilazuarda2024towards}. 
This ``inverse problem'' is compounded by the fact that AI researchers typically lack training in anthropology, regional expertise, or other disciplines focused on cultural study.  Furthermore,  major AI firms do not maintain standardized metrics for concepts like cultural competency \citep{pawar2025culture}, 
leaving existing benchmarks primarily at the research stage.  
As GenAI confronts users with the ability to represent--or misrepresent--culturally sensitive imagery at scale, several urgent questions emerge: 
How should developers located in the Global North understand ``culture'' for a global audience? What aspects of culture are sensitive in a GenAI context?  

To begin to mitigate this asymmetrical situation, we conducted a multi-country survey across 13 nations (\textit{n}=5,629). This study provides empirical evidence on which categories of \textit{culture} global populations find particularly important and sensitive. Grounded in UNESCO's (n.d., 2025) frameworks around cultural heritage, our analysis is guided by the following research questions: 

\begin{enumerate}
    \item What elements of identity (e.g., traditions, language, religion) do people in diverse global regions consider most central to their culture?
   \item How does user familiarity with generative AI shape their perceptions of potential harms, such as misrepresentation and stereotyping, in cultural depictions?
    \item Which cultural topics and practices are perceived as most vulnerable to harmful misrepresentation by generative AI?
\end{enumerate}
Our findings highlight geographic variations in how ``culture'' is understood and establish critical ``redlines'' for GenAI representation, particularly regarding religion and tradition. 

\section{Literature Review}
\subsection{What is culture/what does it mean}
The ability of GenAI systems to navigate cultural nuance and difference remains a persistent challenge for the field \citep{pawar2025culture, alenichev2025we}. 
A fundamental obstacle  to progress is the lack of grounded understandings of ``culture'' against which models can be reliably benchmarked. 
In academic discourse, the meaning of ``culture'' is often contingent upon the specific object of study. In linguistics, for instance, it manifests as communicative cues; in psychology, it describes interpersonal mental processes; and in anthropology, it encompasses shared patterns of behavior.

Developing culturally aligned GenAI is thus particularly difficult because ``culture'' is an essentially contested concept \citep{gallie1955essentially}. 
Indeed, there have likely been more anthropological definitions of culture than there have been anthropologists \citep{monaghan2000social}. 
While culture is fundamentally a \textit{shared} set of beliefs and social practices,  individuals often maintain idiosyncratic interpretations of their own cultural identity \citep{marti2023latent, friedman1994cultural}. 
Despite this variation, social scientists generally define culture as shared patterns of learned behavior that vary geographically, and along dimensions of social identity, such as race, class, gender, or disability \citep{bernard2017research, herzfeld2014cultural, mamadouh1999grid}. 

Complementary to these behavioral views, culture functions as an epistemic lens through which people interpret the world  \citep{harris2001rise, Purvis1993DiscourseID}.
These lenses are often shaped by structural hierarchies; most societies, for instance, exhibit forms of male-domination and ethno-racial stratification that influence social practices \citep{mcclaurin2025black, lewin2006feminist}.
Furthermore, culture is dynamic rather than static \citep{bernard2017research}. 
and is never truly isolated \citep{friedman1994cultural}. 
Throughout history, nomadic movement and ecological adaptation have placed cultures in constant contact through blending, merging, or conquering \citep{graeber2021dawn}. 
While globalization has dominated recent decades, a rising nationalism suggests that local cultures remain resistant to total absorption into a global capitalist economy \citep{appiah2023ethics}. 
The role GenAI will play in this shifting landscape remains an open question.   

Following these considerations, we adopt a pragmatic framework for our empirical survey grounded in the UNESCO Thesaurus (n.d.) (cultural and natural heritage; visual arts and crafts; books and press; audio-visual and interactive media; sports and recreation) and the five domains of Intangible Cultural Heritage listed by UNESCO (2025) (Traditional craftmanship and design; knowledge and practices on nature and the universe; performing arts; social practices, rituals, and festive events; oral traditions and expressions). This aligns our study with internationally recognized concepts.

\subsection{Technology and Culture}
Users and developers hold ``imaginaries''--envisioned futures--of what technology can and should do for within cultural life. We adopt the view of \textit{sociotechnical imaginaries} \citep{jasanoff2009nuclear, jasanoff2015sti}
as vital cultural resources that produce ``systems of meaning that enable collective interpretations of social reality.'' 
These imaginaries are essential for understanding which aspects or dimensions of culture people find most sensitive in GenAI depictions.

Through this lens, GenAI serves as a technological cultural artifact, that is, a product of specific engineering choices rooted in Western concepts of rationality, even colonialism \citep{alenichev2025we}. 
GenAI models match high-dimensional numerical vector representations multimodally (e.g., text-to-image or vice versa); these numerical representations are learned from the inherently cultural  datasets that these models are trained on. 
The vector representations form a ``latent space'' where words and images with similar meanings are positioned near one another. As this ``nearness''  is learned from  training data, the way GenAI represents culture is already heavily biased by its source material. The challenge lies in steering this technology to represent diverse cultures authentically, without violating established cultural norms.

\subsection{Cultural Representation in Data}
GenAI frequently replicates the `Western Educated Industrialized Rich Democratic' (WEIRD) bias common in psychology, a field where 80\% of participants come from populations that comprise just 12 \% of the global total \citep{henrich2010weird}.
While techniques like Reinforcement Learning from Human Feedback (RLHF) provide some behavioral control over model output, aligning GenAI with pluralistic global values remains difficult \citep{khan2025randomness} \citep{smart2024socially}. 
Current alignment methods often reduce cultural alignment to simple binary-choice ``preference'' queries that lack robust social science validation. 
Attempts to expand the diversity of cultural viewpoints, such as PRISM \citep{kirk2024prism} 
or NormAd \citep{rao2024normad}, 
typically focus on predefined aspects of culture, failing to capture how communities define culture for themselves. 

In the absence of large-scale ethnographic research, surveys offer a useful alternative for capturing population-level perceptions. 
Survey-based assessments like the World Values Survey (WVS) \citep{inglehart2000world} 
and Global Opinion QA \citep{durmus2023towards}  
track the economic, political and social consequences of changes in mass belief systems. However, recent algorithmic assessments using the WVS have revealed that  Western biases persist in LLMs despite these efforts \citep{benkler2023assessing}. 
To prevent cultural harm, it is necessary to build an empirical understanding of how global populations view their own  ``redlines'' for GenAI representation. 

This paper explores how global perspectives relate to expectations of GenAI. 
Unlike traditional efforts to produce inclusive datasets, which often face inherent limitations \citep{qadri2025confusing}, 
our approach prioritizes surfacing the ``pain points'' and values how communities identify themselves. By doing so, we establish a grounded baseline for the controls necessary to avoid cultural misrepresentation in GenAI systems. 

\section{Methodology}
The research team fielded a cross-sectional survey on culture and GenAI between September 2024 and January 2025. Recruitment included a nationally representative sample of 5,629 respondents from 13 countries: Brazil, Cameroon, France, Germany, India, Indonesia, Italy, Japan, Mexico, Nigeria, South Korea, the United Arab Emirates, and the United States. 
The selected countries represent a wide range of geographic regions and  variance in cultural dimensions \citep{muthukrishna2020}.  

\subsection{Participants}
To qualify for the study, participants were required to be at least 18 years of age, current residents of a target country, and able to complete the survey in the local language.
We aimed for a minimum of 300 adults (18+) per country; However, Cameroon's sample size was smaller (\textit{n}=56) due to its smaller population and specific regional recruiting constraints. The final sample size totaled 5,629 respondents. 
We set quotas for gender at  50\% male and 50\% female for each country. In the final aggregate sample, women accounted for 50.9\% of respondents and men for 49.8\%. Respondents identifying as “non-binary/third gender,” “prefer to self describe,” or who indicated “prefer not to answer” comprised the remaining 0.3\%. We aimed to recruit evenly across five age ranges (18-24, 25-34, 35-44, 45-55, 55+). While most countries adhered to these quotas, some distributions varied due to local recruitment challenges.  
Table \ref{tab:demographics} reports demographic breakdowns for all 13 countries.

\begin{table}[h]
    \centering
    \caption{\textbf{Participant Demographics}}
    \footnotesize
    \label{tab:demographics}
    \begin{tabularx}{\textwidth}{l *{7}{>{\centering\arraybackslash}X}}
        \toprule
        \textit{Country} & \textbf{Brazil} & \textbf{Camer.} & \textbf{France} & \textbf{Germ.} & \textbf{India} & \textbf{Indon.} & \textbf{Italy} \\
        \midrule
        \textit{Survey language} & Portug. & French & French & German & English & Bahasa & Italian \\
        \midrule
        \textit{Respondent count} & 464 & 56 & 510 & 504 & 514 & 484 & 529 \\
        \midrule
        \multicolumn{8}{l}{\textbf{Gender}} \\
        \hspace*{2mm} Man & 49.6\% & 55.4\% & 46.5\% & 48.8\% & 48.4\% & 49.4\% & 50.1\% \\
        \hspace*{2mm} Woman & 50.2\% & 44.6\% & 53.3\% & 50.8\% & 51.36\% & 60.6\% & 49.5\% \\
        \hspace*{2mm} Nonbinary/Third gender & 0.2\% & - & - & - & 0.2\% & - & 0.2\% \\
        \hspace*{2mm} Prefer not to answer & - & - & - & 0.4\% & 0.2\% & - & - \\
        \hspace*{2mm} Prefer to self-describe & - & - & - & - & - & - & 0.2\% \\
        \midrule
        \multicolumn{8}{l}{\textbf{Age}} \\
        \hspace*{2mm} 18--24 & 40.0\% & 23.21\% & 17.8\% & 19.8\% & 29.7\% & 33.3\% & 21.2\% \\ 
        \hspace*{2mm} 25--34 & 20.0\% & 25.0\% & 18.7\% & 17.7\% & 40.5\% & - & 17.4\%\\ 
        \hspace*{2mm} 35--44 & - & 25.0\% & 17.3\% & 18.8\% & 13.5\% & 66.7\% &  17.8\%\\
        \hspace*{2mm} 45--54 & 40.0\% & 21.43\% & 21.7\% & 20.8\% & 10.8\% & - & 19.1\% \\
        \hspace*{2mm} 55--64 & - & 3.57\% & 13.7\% & 12.5\% & 2.7\% & - & 15.9\% \\ 
        \hspace*{2mm} 65--74 & - & - & 9.4\% & 9.8\% & - & - & 6.8\% \\ 
        \hspace*{2mm} 75+ & - & 1.70\% & 1.4\% & 0.6\% & 2.7\% & - &1.8\%  \\ 
        \hspace*{2mm} Prefer not to answer & - & - & - & 0.1\% & - & - & 0.1\% \\ %
        \toprule
       \textit{Country} & \textbf{Japan} & \textbf{Mexico} & \textbf{Nigeria} & \textbf{S.Korea} & \textbf{UAE} & \textbf{US} & \textbf{\phantom{Country}} \\ 
        \midrule
       \textit{Survey language} & Japanese & Spanish & English & Korean & Arabic & English & \\ 
        \midrule
        \textit{Respondent count} & 320 & 302 & 410 & 505 & 533 & 498 & \\
        \midrule
        \multicolumn{8}{l}{\textbf{Gender}} \\
        \hspace*{2mm} Man & 43.8\% & 49.0\% & 48.8\% & 49.7\% & 51.2\% & 47.8\% & \\
        \hspace*{2mm} Woman & 55.3\% & 50.3\% & 51.0\% & 49.9\% & 48.6\% & 52.0\% & \\ 
        \hspace*{2mm} Nonbinary/Third gender & 0.3\% & 0.7\% & 0.2\% & 0.2\% & - & - & \\ 
        \hspace*{2mm} Prefer not to answer & 0.3\% & - & - & 0.2\% & 0.2\% & - & \\ 
        \hspace*{2mm} Prefer to self-describe & 0.3\% & - & - & - & - & 0.2\% & \\ 
        \midrule
        \multicolumn{8}{l}{\textbf{Age}} \\
        \hspace*{2mm} 18--24 & 26.7\% & 20.0\% & 50.0\% & 17.9\% & 20.9\% & 55.2\% & \\ 
        \hspace*{2mm} 25--34 & 53.3\% & 19.87\% & 50.0\% & 19.8\% & 18.0\% & 20.7\% & \\ 
        \hspace*{2mm} 35--44 & 6.7\% & 19.54\% & - & 19.5\% & 15.8\% & 20.7\% & \\ 
        \hspace*{2mm} 45--54 & 13.3\% & 19.87\% & - & 20.6\% & 23.5\% & - & \\
        \hspace*{2mm} 55--64 & - & 18.21\% & - & 15.5\% & 20.4\% & 3.4\% & \\ 
        \hspace*{2mm} 65--74 & - & 2.32\% & - & 6.3\% & 1.5\% & - & \\ 
        \hspace*{2mm} 75+ & - & 0.33\% & - & 0.5\% & - & - & \\ 
        \hspace*{2mm} Prefer not to answer & - & - & - & - & - & - & \\ 
        \bottomrule
    \end{tabularx}
\end{table}

\subsection{Study Design}
As no validated instrument existed to assess public perceptions of AI-related cultural harms across diverse national contexts, survey items were developed through a review of theoretical and empirical literature on cultural values, AI ethics, and GenAI and consultation with the research team's expertise in linguistics, anthropology, Human-Computer Interaction, and sociology. 
The instrument employs both emic (culture-specific) perspectives through open-ended items and etic (cross-culturally comparable) perspectives through standardized closed-ended items.
Four subject matter experts in cultural studies, responsible AI governance, and survey methodology reviewed the item pool for content validity, clarity, and cultural appropriateness, and provided qualitative feedback. 
Items were revised based on feedback, resulting in 24 items retained for pilot testing. The revised instrument was piloted (\textit{n} = 287) to assess comprehension and cultural appropriateness.


\subsubsection{Survey Structure}
The current analysis focuses on 8 items from the larger 24-item survey. Prospective participants first received an explanation of the survey's focus on culture and GenAI followed by screener questions to enable quota sampling. After qualifying, respondents completed a consent  form, and received locally relevant compensation for their participation.

\begin{itemize}
    \item \textbf{Understanding of Culture.} An open-ended question asked respondents to decribe, in their own words, ``What is your understanding of the word culture?''  This question was placed near the beginning of the survey to avoid priming respondents with the closed-ended cultural categories.  
    \item \textbf{Social Identities.} Respondents indicated which three of 19 social identities (gender, race/color, ethnicity/tribe, nationality, sexual orientation, caste, religion or tradition, health status, disability, family, education, immigration status, military or veteran status, area where I live, occupation or profession, political affiliation, socioeconomic status, asylum or refugee status, and other) they consider important when thinking about culture. 
    \item \textbf{Important Aspects of Culture.}  To understand the salient components of respondents' cultural experiences, they were asked in an open-ended question to describe what kinds of artifacts, practices or other attributes they consider most important. 
    \item \textbf{GenAI Familiarity.} Familiarity with GenAI was assessed using a single-item measure where respondents indicated ``How familiar are you with generative AI?'' on a four point Likert scale. 
    \item \textbf{Cultural Redlines.} To gauge attitudes towards representation boundaries, respondents were first asked, ``Are there any aspects of your culture that you believe should NEVER be represented in generative AI tools?'' (no, yes). 
    \item \textbf{Prohibitive Representation.}  Respondents who indicated `yes' to the cultural redlines question were then asked to elaborate in an open-text field describing ``What aspects of your culture should NEVER be represented?'' 
    \item \textbf{Approach for Ambiguous Cultural Depiction.} Participants were asked in instances ``When generative AI is unsure about how to depict an aspect of culture, which approach is most appropriate?''(Provide a disclaimer explaining the uncertainty, offer multiple variations and let the user choose, decline to generate content related to that aspect, Other) 
    \item \textbf{Sensitivity Ranking Cultural Categories.} Participants were asked to rank-order a provided list of cultural categories based on their perceived sensitivity for GenAI to depict or describe, from most sensitive to least sensitive. Response options were drawn from the UNESCO Thesaurus (n.d.) 
    (cultural and natural heritage;  visual arts and crafts; books and press; audio-visual and interactive media; sports and recreation)  and the five domains of Intangible Cultural Heritage listed by UNESCO (2025) (Traditional craftsmanship and design; knowledge and practices on nature and the universe; performing arts; social practices, rituals, and festive events; oral traditions and expressions) \citep{unesco2025ich} 
    to align to internationally recognized concepts. 
    \item \textbf{Sensitivity Ranking Social Identities.} In a separate question, participants were asked to rate social identities according to a five point Likert scale based on perceived sensitivity for GenAI to depict or describe, from most sensitive to least sensitive. The list of 19 social identities was the same list from the above question on their importance when thinking about culture. 
\end{itemize}
   


\subsection{Survey Translation}
The survey was developed in English and translated into locally dominant languages for nine countries: Brazil, France, Germany, Indonesia, Italy, Japan, Mexico, South Korea, and UAE. In India and Nigeria, the survey was administered in English to minimize technological barriers for respondents in these multilingual contexts. 

\subsection{Data Analysis}
\subsubsection{Quantitative Analysis}
Quantitative analyses were conducted using Python, specifically Pandas, Numpy, Seaborn and Matplotlib. The data from 13 countries across two waves were initially weighted to reflect the local population characteristics within each country and wave. We then applied a secondary post-stratification weighting step, based on World Bank (2024) data, to ensure  the combined sample reflects the true population size proportions across the 13 surveyed countries.
This paper focuses on descriptive statistics and aggregate measures of salient cultural categories with respect to GenAI; we do not report  within-country demographic comparisons (e.g., men and women, age ranges). 


\subsubsection{Qualitative Analysis}
We conducted an inductive  thematic analysis  \citep{terry2017thematic} \citep{fereday2006demonstrating}
of the open-ended responses from each country regarding respondents' understanding of culture, important aspects of culture, and prohibitive representation. 
Responses in languages other than English were automatically translated and reviewed by members of the research team for accuracy prior to analysis.  
Qualitative themes are integrated with quantitative findings to contextualize cross-cultural patterns and illustrate quantitative differences in cultural centrality and harm perceptions across countries.

\section{Results}
Analysis of the 13-country survey reveals that while global understandings of culture remain rooted in heritage and tradition, the introduction of GenAI shifts sensitivities toward functional and professional identities.

\subsection{Understandings of Culture, Social Identities, \& Important Aspects of Culture}
\label{understanding}
To establish a baseline for cultural representation, we analyzed responses concerning the general understanding of culture, alongside their prioritization of 19 social identities. 
Broadly, when asked to describe ``culture'' in their own words, respondents consistently defined it as a multifaceted composite of language, traditions, beliefs, and artistic expressions. However, regional nuances emerged (see Table \ref{tab:example-responses-word}). European respondents tended to emphasize tangible artifacts, such as \textit{``music''}, \textit{``literature,''} and \textit{``history''}, whereas Asian respondents focused on collective heritage, frequently citing \textit{``ancestors,''} \textit{``nation,''} and \textit{``country.''} Sub-Saharan African respondents centered their definitions on \textit{``collections of attitudes''} and as a \textit{``total way of life.''}

\begin{figure}[H]
    \centering
    \includegraphics[width=1\linewidth]{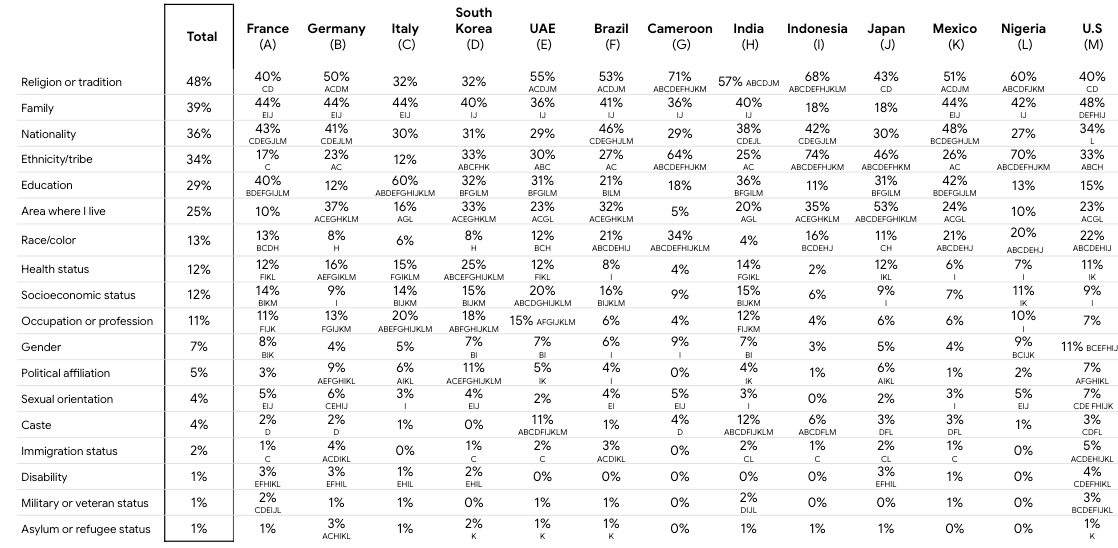}
    \caption{Distribution of social identities selected as ``important'' by country shown as overall and individual percentages. A/B/C/D/E/F/G/H/I/J/K/L/M represent statistical significance at the 95\% confidence level.}
    \label{fig:country-aspects}
\end{figure}

Respondents were also asked to select the three social identities they consider most important to their culture. A global consensus emerged around  ascribed characteristics (see Figure \ref{fig:country-aspects}).  
\textit{Religion and Tradition} was consistently identified as the primary anchor of cultural identity (48\%), ranking in the top three across every country. 
\textit{Nationality} (36\%) and \textit{Ethnicity/tribe} (34\%) were similarly ranked as highly important, with \textit{Ethnicity/tribe} being highest in Nigeria (70\%). 
Conversely, identities associated with individual life experience ranked lower, including \textit{Socioeconomic Status} (12\%), \textit{Health Status} (12\%),  and \textit{Occupation or Profession} (11\%). Regional outliers further illustrated the localized nature of identity; for instance, ``Caste'' was cited by 12\% of respondents in India and 11\% in UAE compared to 4\% or less elsewhere. 

Regarding cultural markers, respondents were asked to share what artifacts, practices, or other attributes they consider most important, identifying landmarks (e.g., \textit{``Lobe waterfalls''} in Cameroon), architecture (e.g., \textit{``Gyeongokygung Palace,''} \textit{``Sungnyemun Gate,''} and \textit{``Bulguksa Temple''} in South Korea) and natural spaces (e.g.,\textit{ ``The Alps''} in France), and food (e.g., \textit{``feijoada''} in Brazil and \textit{``harees'' }in UAE) as essential. Regional variations were prominent: African respondents frequently cited arts (e.g., \textit{``pottery,''} \textit{``Nok culture,''} and `\textit{`textiles''} in Nigeria; and \textit{``monikim''} and \textit{``masques''} in Cameroon), cultural events (e.g., \textit{``New Yam,''} \textit{``Osun Osogbo,''} and \textit{``Masquerade''} in Nigeria), and language (e.g., \textit{``Igbo''} and \textit{``Yoruba''} in Nigeria), while European respondents emphasized iconic architecture (e.g., \textit{``Brandenburg Gate,''} \textit{``Cologne Cathedral,''} \textit{``Neuschwanstein Castle,''} and the \textit{``Berlin Wall''} in Germany; \textit{``the Eiffel Tower,''} \textit{``Arc de Triomphe,''} and the \textit{``Louvre''} in France) and fine arts (e.g., \textit{``Michelangelo's David,''} \textit{``The Divine Comedy'',} and \textit{``The Renaissance era''} in Italy) (see Table \ref{tab:example-responses-artifacts} for exemplar quotes).

\begin{table*}
    \centering
    \footnotesize
    \caption{\label{tab:example-responses-word}
    Example responses from participants' and the corresponding country in the open-ended question: ``What is your understanding of the word culture.''}
    \begin{tabularx}{\textwidth}{l *{7}{>{\arraybackslash}X}}
        \toprule
        \textbf{Country} & \textbf{Response} & \textbf{English Translation} \\
        \midrule
        Brazil & ``tradicao de um pais traduzido em na sociedade atravez da musica da arte dos custumes'' & \textit{Tradition of a country translated into society through music, arts, and customs} \\
        \cmidrule(l){2-3}
        Cameroon & ``Ensemble  des attitudes transmises de génération en génération sans questionnements'' & \textit{Collection of attitudes transmitted from generation to generation without questioning} \\
        \cmidrule(l){2-3}
        France & ``En fonction de ses origines c’est sa manière de vivre'' & \textit{According to one's origins it's the way of life} \\
        \cmidrule(l){2-3}
        Germany & ``Eine Zusammenfassung von Kunst, Musik und ähnlichen Angeboten, die einem gewissen Rahmen als Nation und Identität untergeordnet wird.'' & \textit{A compilation of art, music, and similar offerings that are subordinated to a certain framework of nation and identity.} \\
        \cmidrule(l){2-3}
        India & ``a group of people who share a common set of beliefs, values, and practices that shape their way of life'' &  — \\
        \cmidrule(l){2-3}
        Indonesia & ``Budaya adalah warisan turun temurun'' & \textit{Culture is heritage} \\
        \cmidrule(l){2-3}
        Italy & ``Per cultura intendo l'insieme di conoscenza, credenze, arte, morale, diritto, costume e tutte quelle capacità e abitudini acquistate da ogni individuo'' & \textit{By culture, I mean the set of knowledge, beliefs, art, morals, law, customs, and all those abilities and habits acquired by each individual.} \\
        \cmidrule(l){2-3}
        Japan & `` その国や地域、物事の歴史が紡がれたときにできた性格や好み '' & The characteristics and preferences that were formed when the history of that country, region, or thing was woven together. \\
        \cmidrule(l){2-3}
        Mexico & ``Es un elemento fundamentalmente en la construcción de la identidad individual y colectiva y se trasmite de generación en generación'' & \textit{It is a fundamental element in the construction of individual and collective identity and is transmitted from generation to generation.} \\
        \cmidrule(l){2-3}
        Nigeria & ``Culture is the total way of life of a group of people in terms of language, dressing, food and so on.'' & — \\
        \cmidrule(l){2-3}
        South Korea & `` 지식, 신앙, 생활, 습관, 법률, 질서 등에 있어 사람과 사람사이에서 만들어진 사고나 태도 그리고 행위의 복합태라고 생각합니다. '' & \textit{I believe it is a complex combination of thoughts, attitudes, and behaviors created between people in areas such as knowledge, faith, lifestyle, habits, laws, and order.} \\
        \cmidrule(l){2-3}
        UAE & \textarabic{``هوية المجتمع واعرافه وتقاليده ومستوى المعرفة الخاص بكل مستوى}'' & \textit{The community's identity, customs, traditions, and the level of knowledge specific to each level.} \\
        \cmidrule(l){2-3}
        USA & ``Where you live, the foods you eat specific to your area, your beliefs, customs, heritage'' & — \\
        \bottomrule
    \end{tabularx}
\end{table*}

\begin{table*}
    \centering
    \footnotesize
    \caption{\label{tab:example-responses-artifacts}
    Example responses from participants' and the corresponding country in the open-ended question: ``What artifacts, practices, or other attributes do you consider most important?''}
    \begin{tabularx}{\textwidth}{l *{7}{>{\arraybackslash}X}}
        \toprule
        \textbf{Country} & \textbf{Response} & \textbf{English Translation} \\
        \midrule
        Brazil & ``O Maracatu do Nordeste, O Carnaval, O Samba, Os Rituais Indígenas, etc'' & \textit{Maracatu from the Northeast, Carnival, Samba, Indigenous rituals, etc.} \\
        \cmidrule(l){2-3}
        Cameroon & ``Le monument de la. Réunification,le palais de l'unité, la chefferie traditionnelle.'' & \textit{The Reunification Monument, the Unity Palace, the traditional chiefdom} \\
        \cmidrule(l){2-3}
        France & ``Lorsque je pense à ma culture, des éléments comme les festivals traditionnels, l’architecture emblématique comme la Tour Eiffel, la gastronomie avec des plats typiques, et les rituels religieux sont des aspects essentiels qui incarnent mon identité culturelle.'' & \textit{When I think about my culture, elements such as traditional festivals, iconic architecture like the Eiffel Tower, gastronomy with typical dishes, and religious rituals are essential aspects that embody my cultural identity.} \\
        \cmidrule(l){2-3}
        Germany & ``Deutsche Dichter und Erfinder, wie Goethe, Heine, Lilienthal usw'' & \textit{German poets and inventors, such as Goethe, Heine, Lilienthal, etc.} \\
        \cmidrule(l){2-3}
        India & ``We do have many temples and churches around, and so the people are very orthodox and traditional around &  \\
        \cmidrule(l){2-3}
        Indonesia & ``Setiap mau selamatan misal hajatan, buat rumah.... Kita selalu menyediakan seperti sesajen untuk berdoa kepada Tuhan juga para Leluhur agar diberikan kemudahan juga kelancaran'' & \textit{Whenever we have a special occasion, such as a celebration or building a house, we always prepare offerings to pray to God and our ancestors for ease and smooth proceedings.} \\
        \cmidrule(l){2-3}
        Italy & ``Arte e architettura : Colosseo, la torre di Pisa, musei vaticani, la Basilicata di San Pietro. poi sicuramente la cucina e la moda'' & \textit{Art and architecture: the Colosseum, the Leaning Tower of Pisa, the Vatican Museums, St. Peter's Basilica. And of course, the cuisine and fashion.} \\
        \cmidrule(l){2-3}
        Japan & ``古くから伝わる物語'' & \textit{Stories passed down from ancient times} \\
        \cmidrule(l){2-3}
        Mexico & ``La torre latinoamericana, festividades de día de muertos, el ángel de la independencia,  los arcos de Querétaro '' & \textit{The Latin American Tower, Day of the Dead festivities, the Angel of Independence, the arches of Querétaro} \\
        \cmidrule(l){2-3}
        Nigeria & ``Culture is the total way of life of a group of people in terms of language, dressing, food and so on.'' & \\
        \cmidrule(l){2-3}
        South Korea & ``경주 불국사 숭례문'' & \textit{Gyeongju Bulguksa Sungnyemun Gate} \\
        \cmidrule(l){2-3}
        UAE & ``\textarabic{الملابس التقليدية لكل بلد تعكس ثقافتهم }'' & \textit{The traditional clothing of each country reflects their culture.} \\
        \cmidrule(l){2-3}
        USA & ``traditional cuisine, family-centered celebrations like holidays and gatherings'' & \\
        \bottomrule
    \end{tabularx}
\end{table*}

\subsection{From Cultural Importance to Sensitivity in GenAI}
\begin{figure}[t!]
    \centering
    \includegraphics[width=1\linewidth]{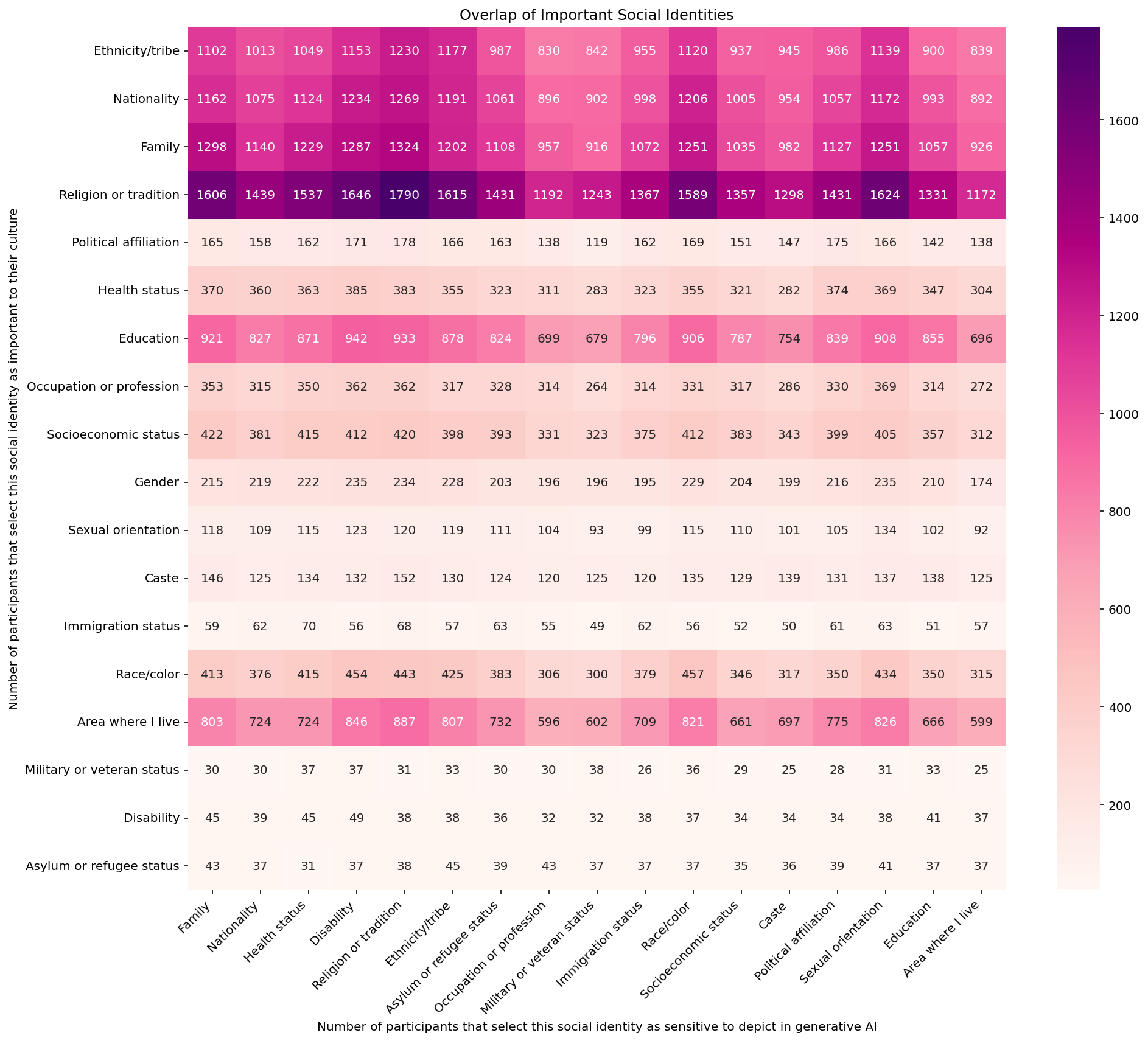}
    \caption{Social identities selected as important and sensitive for for all countries. This shows the relationship between the importance (Y-axis) and perceived GenAI sensitivity (X-axis). Darker shades indicate a high density of respondents who categorize a social identity as both culturally salient and technologically sensitive. Notably \textit{religion or tradition} stands out as a salient identity}
    \label{fig:overall-heatmap}
\end{figure}

To analyze shifts in perception when culture is framed in a GenAI context, we compared the cultural  \textbf{importance} rankings (see Section \ref{understanding}) with perceived GenAI sensitivity. While every attribute saw a lower frequency in sensitivity counts compared to importance counts, the ratio of ``agreement''--the frequency with which an aspect marked as important was also deemed sensitive--varied significantly. 
We utilized frequency co-occurrence heatmaps to visualize the relationship between the  \textbf{importance} (Y-axis) and perceived GenAI \textbf{sensitivity} (X-axis). Darker shades indicate a high density of respondents who categorize a social identity as both culturally salient and technologically sensitive. See Figure \ref{fig:overall-heatmap} for a summary.

\textbf{Stability in High-Centrality Markers.} \textit{Religion or Tradition} exhibited the highest level of stability and density across importance and sensitivity. It was identified as important and  sensitive (65.6\%). For those who viewed \textit{Religion or Tradition} as important, significant numbers also flagged \textit{Nationality} (52.3\%) and \textit{Ethnicity/Tribe} (61.1\%) as sensitive, see Table \ref{tab:sensitivity_agreement} for overall agreement rates. 

\begin{table}[ht]
    \centering
    \footnotesize
    \caption{Agreement Rates on Importance and Sensitivity Across Identity Categories for All Countries Combined.}
    \label{tab:sensitivity_agreement}
    \begin{tabular}{l S[table-format=5.0] S[table-format=5.0] S[table-format=1.3]}
        \toprule
        \textbf{Category} & {\textbf{Importance Count}} & {\textbf{Sensitive Count}} & {\textbf{Agreement Rate}} \\
        \midrule
        Military or veteran status & 51 & 38 & 0.745 \\
        Religion or tradition & 2729 & 1790 & 0.656 \\
        Disability & 75 & 49 & 0.653 \\
        Caste & 215 & 139 & 0.647 \\
        Sexual orientation & 208 & 134 & 0.644 \\
        Political affiliation & 279 & 175 & 0.627 \\
        Race/color & 746 & 457 & 0.613 \\
        Ethnicity/tribe & 1924 & 1177 & 0.612 \\
        Family & 2185 & 1298 & 0.594 \\
        Asylum or refugee status & 66 & 39 & 0.591 \\
        Immigration status & 110 & 62 & 0.564 \\
        Socioeconomic status & 696 & 383 & 0.550 \\
        Health status & 662 & 363 & 0.548 \\
        Education & 1606 & 855 & 0.532 \\
        Nationality & 2038 & 1075 & 0.527 \\
        Occupation or profession & 621 & 314 & 0.506 \\
        Area where I live & 1425 & 599 & 0.420 \\
        Gender & 36 & 0 & 0.000 \\
        \bottomrule
    \end{tabular}
\end{table}

\textbf{Perceptual Pivots and Localized Sensitivities.} Notably, the analysis identified ``low-importance, high-sensitivity'' categories where identities were flagged as disproportionately sensitive in a GenAI context. Although only 4\% of overall respondents ranked \textit{Caste} as a top cultural marker (Figure \ref{fig:country-aspects}), it had one of the highest rates for sensitivity at 64.6\% agreement. \textit{Military or Veteran Status} saw the highest overall agreement rate (74.5\% agreement), suggesting that when this identity is salient to a respondent, it is almost universally viewed as a sensitive ``redline'' for GenAI representation. Conversely, some high-importance markers showed lower sensitivity agreement, such as \textit{Area where I live} (42\% agreement). 
These findings highlight that while certain identity markers like \textit{Religion or Tradition} maintain a stable cross-national profile, the introduction of GenAI spurs unique concerns. Specifically, identities that are are ascribed or inherited, such as \textit{Ethnicity/tribe} (61.1\% agreement) and \textit{Caste} (64.6\% agreement), consistently generate higher sensitivity than ``achieved'' or life-experience identities like \textit{Occupation or Profession} (50.5\% agreement) or \textit{Education} (53.2\% agreement). 

\begin{figure}[H]
    \centering
    \includegraphics[width=0.85\linewidth]{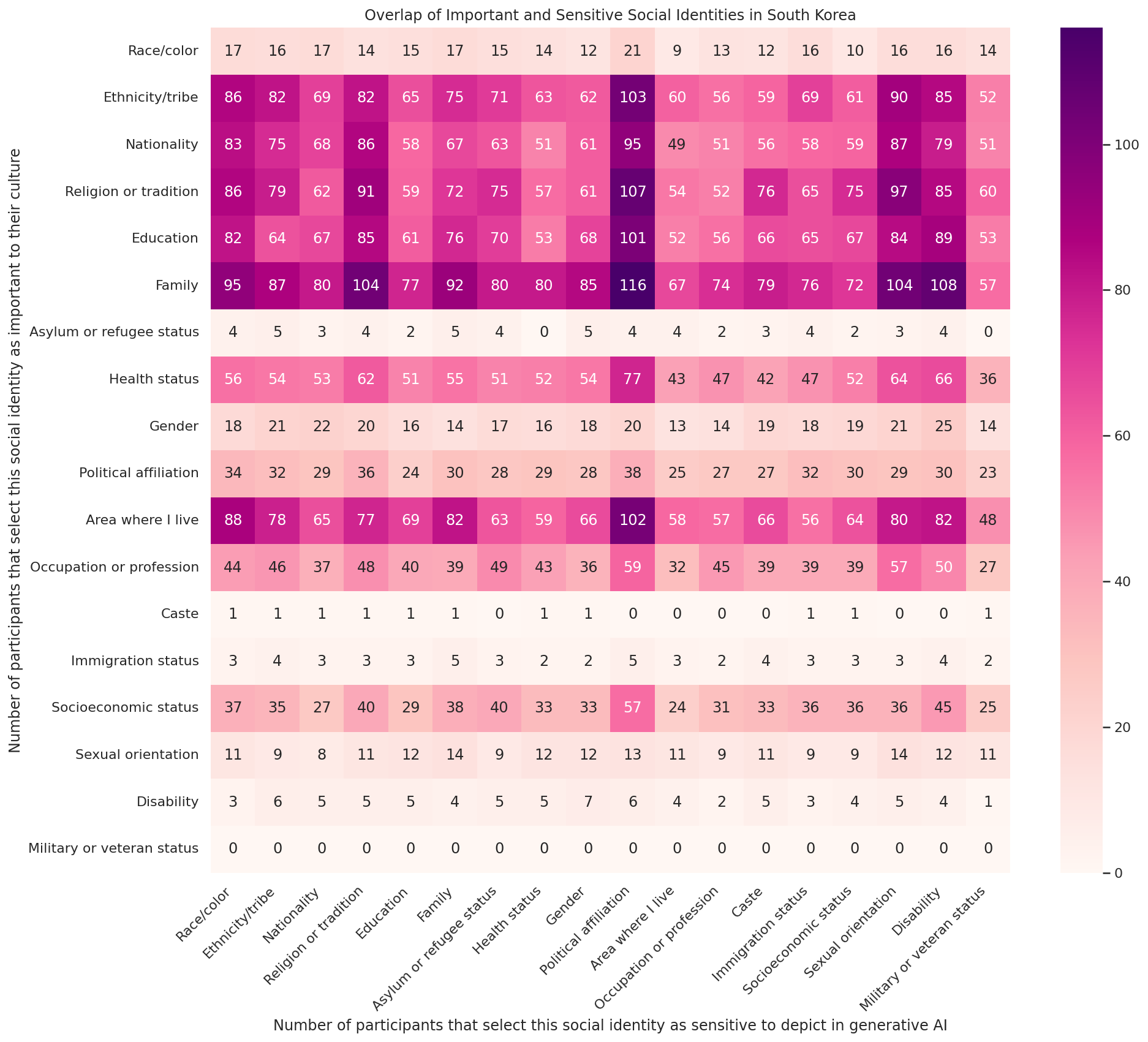}
    \caption{Social identities selected as important and sensitive for South Korea. This shows the relationship between the importance (Y-axis) and perceived GenAI sensitivity (X-axis). Darker shades indicate a high density of respondents who categorize a social identity as both culturally salient and technologically sensitive. Notably \textit{family} stands out as a salient identity and \textit{health status} is unique to this country.}
    \label{fig:southkorea-heatmap}
\end{figure}

\begin{figure}[H]
    \centering
    \includegraphics[width=0.85\linewidth]{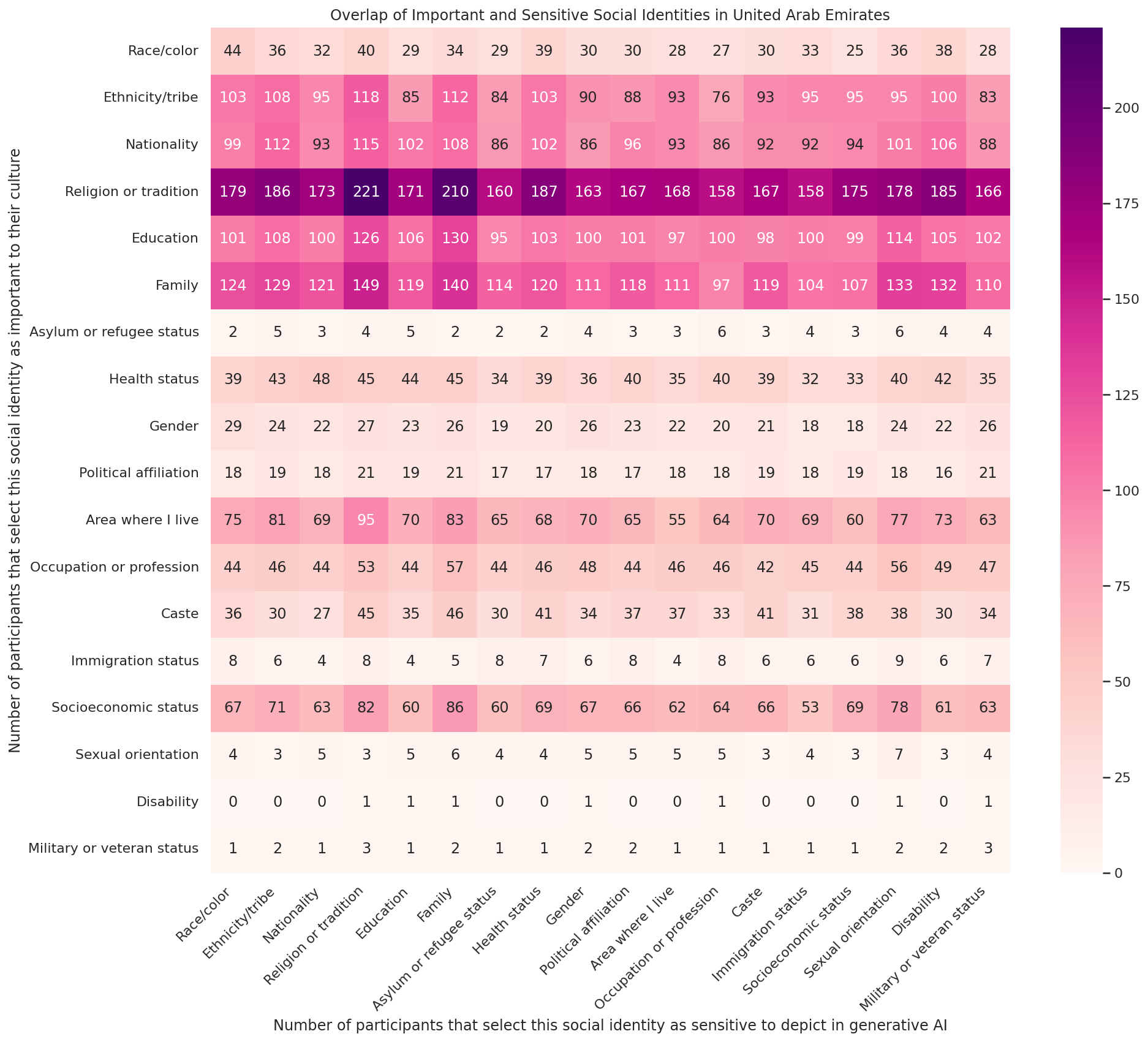}
    \caption{Social identities selected as important and sensitive for UAE. This shows the relationship between the importance (Y-axis) and perceived GenAI sensitivity (X-axis). Darker shades indicate a high density of respondents who categorize a social identity as both culturally salient and technologically sensitive. Notably \textit{religion or tradition} stands out as a salient identity and \textit{socioeconomic status} is unique to this country.}
    \label{fig:uae-heatmap}
\end{figure}

Sensitivities appear localized. For instance, in South Korea, significant shifts occurred toward \textit{Health Status}, \textit{Area where I live}, and \textit{Occupation or Profession} (see Figure \ref{fig:southkorea-heatmap}). Conversely,  the UAE reflected the global trend of stability, with \textit{Religion or Tradition} remaining the dominant identity for both importance and sensitivity (see  Figure \ref{fig:uae-heatmap}). These findings highlight that while certain identity markers like religion maintain a stable cross-national profile, the introduction of GenAI spurs unique regional concerns regarding health, location, and professional status.

\subsection{Familiarity with Generative AI and Expectations}
Broadly, most respondents are familiar with GenAI, with 68\% of all respondents reporting at least some familiarity with the technology. Among those aware, 59\% have actively used GenAI to generate content. Familiarity is most pronounced in South Korea and the UAE, likely driven by  rapid adoption of GenAI, supported by strong government initiatives and significant  digital infrastructure investments. Conversely, respondents in France and Cameroon reported lowest levels of familiarity. 
Detailed country-level results, including 95\% confidence intervals (CIs), are provided in Table \ref{tab:genAIfamiliarity}.

\begin{table}[t!]
    \centering
    \footnotesize
    \caption{Familiarity with GenAI Content, by Country}
    \label{tab:genAIfamiliarity}
    \begin{tabular}{llll}
    \toprule
    \textbf{Region} & \textbf{GenAI Familiarity Level} \% & \textbf{95\%} & \textbf{CI} \\
    \midrule
    \multirow{4}{*}{Brazil} & I've never heard of it until now & 8\% & [5.82-10.81] \\
    & I've only heard about it in passing & 17\% & [14.24-21.15] \\
    & I know a little bit about it & 43\% & [37.75-46.69] \\
    & I'm quite familiar & 32\% & [27.56-35.98] \\
    \hline
    \multirow{4}{*}{Cameroon} & I've never heard of it until now & 25\% & [14.88-37.75] \\
    & I've only heard about it in passing & 34\% & [21.10-45.57] \\
    & I know a little bit about it & 25\% & [13.39-35.74] \\
    & I'm quite familiar & 16\% & [6.32-25.26] \\
    \hline
    \multirow{4}{*}{France} & I've never heard of it until now & 14\% & [11.01-17.01] \\
    & I've only heard about it in passing & 40\% & [35.46-43.92] \\
    & I know a little bit about it & 29\% & [25.44-33.32] \\
    & I'm quite familiar & 17\% & [13.68-20.17] \\
    \hline
    \multirow{4}{*}{Germany} & I've never heard of it until now & 5\% & [3.23-7.09] \\
    & I've only heard about it in passing & 32\% & [27.49-35.60] \\
    & I know a little bit about it & 40\% & [36.19-44.76] \\
    & I'm quite familiar & 23\% & [19.15-26.48] \\
    \hline
    \multirow{4}{*}{India} & I've never heard of it until now & 4\% & [2.05-5.30] \\
    & I've only heard about it in passing & 10\% & [7.81-13.08] \\
    & I know a little bit about it & 38\% & [33.54-41.90] \\
    & I'm quite familiar & 48\% & [43.86-52.47] \\
    \hline
    \multirow{4}{*}{Indonesia} & I've never heard of it until now & 5\% & [3.63-7.71] \\
    & I've only heard about it in passing & 23\% & [19.17-26.58] \\
    & I know a little bit about it & 39\% & [34.37-42.96] \\
    & I'm quite familiar & 33\% & [28.65-36.93] \\
    \hline
    \multirow{4}{*}{Italy} & I've never heard of it until now & 9\% & [6.59-11.46] \\
    & I've only heard about it in passing & 32\% & [28.17-36.11] \\
    & I know a little bit about it & 52\% & [34.40-42.67] \\
    & I'm quite familiar & 19\% & [16.88-23.72] \\
    \hline
    \multirow{4}{*}{Japan} & I've never heard of it until now & 3\% & [1.44-5.42] \\
    & I've only heard about it in passing & 50\% & [44.69-55.63] \\
    & I know a little bit about it & 38\% & [32.09-42.68] \\
    & I'm quite familiar & 9\% & [5.90-12.17] \\
    \hline
    \multirow{4}{*}{Mexico} & I've never heard of it until now & 9\% & [5.70-12.12] \\
    & I've only heard about it in passing & 25\% & [20.20-29.96] \\
    & I know a little bit about it & 45\% & [36.61-50.82] \\
    & I'm quite familiar & 21\% & [16.22-25.36] \\
    \hline
    \multirow{4}{*}{Nigeria} & I've never heard of it until now & 6\% & [3.53-8.01] \\
    & I've only heard about it in passing & 11\% & [8.26-14.34] \\
    & I know a little bit about it & 35\% & [30.98-40.18] \\
    & I'm quite familiar & 48\% & [42.56-52.15] \\
    \hline
    \multirow{4}{*}{South Korea} & I've never heard of it until now & 3\% & [1.19-3.96] \\
    & I've only heard about it in passing & 26\% & [22.31-29.97] \\
    & I know a little bit about it & 52\% & [47.72-56.44] \\
    & I'm quite familiar & 19\% & [15.77-22.64] \\
    \hline
    \multirow{4}{*}{UAE} & I've never heard of it until now & 6\% & [3.92-7.89] \\
    & I've only heard about it in passing & 15\% & [12.11-18.15] \\
    & I know a little bit about it & 43\% & [39.37-47.72] \\
    & I'm quite familiar & 36\% & [31.40-39.45] \\
    \hline
    \multirow{4}{*}{United States} & I've never heard of it until now & 5\% & [3.57-7.59] \\
    & I've only heard about it in passing & 27\% & [22.63-30.35] \\
    & I know a little bit about it & 44\% & [39.68-48.37] \\
    & I'm quite familiar & 24\% & [20.17-27.64] \\
    \bottomrule
    \end{tabular}
\end{table}

\begin{figure}
    \centering
    \includegraphics[width=1\linewidth]{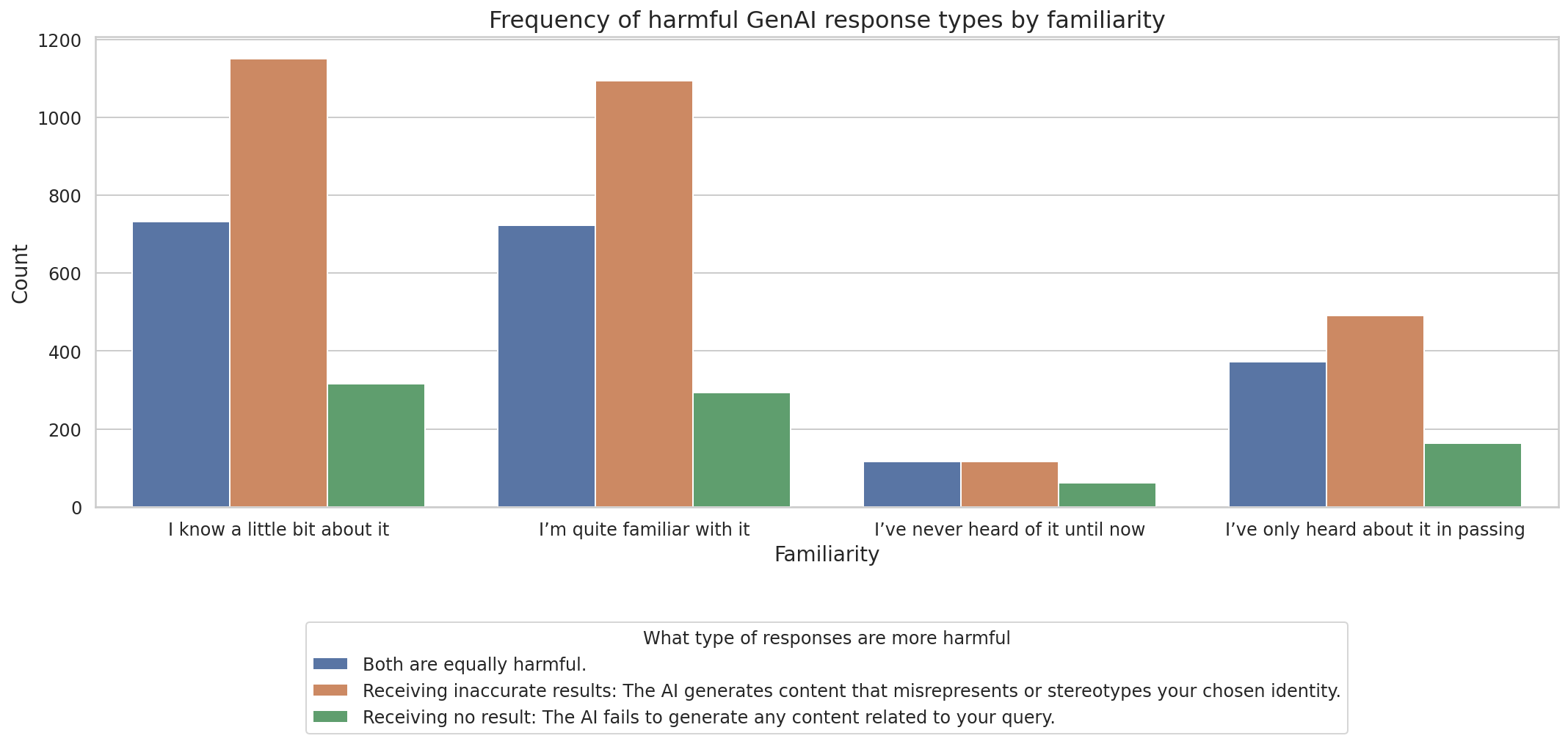}
    \caption{Perception of GenAI harms by user familiarity. Distribution of responses identifying specific GenAI outputs as harmful, categorized by the participant's self-reported familiarity with the technology.}
    \label{fig:familarity-harms}
\end{figure}

Despite high awareness, respondents' opinions regarding GenAI’s ability to represent  identity and culture remains mixed, with approximately half of the respondents expressing ambivalence. Qualitative responses suggest a desire to exclude artistic expression and sacred rituals from AI generation, with many respondents arguing that GenAI lacks the intrinsic human emotions and creativity required for authentic cultural representation. Quantitatively, ``inaccurate information'' is perceived as the primary threat to cultural representation, as illustrated in Figure \ref{fig:familarity-harms}. This concern remains consistent regardless of  a participant's familiarity with GenAI. 

\subsection{Cultural Redlines in GenAI Results}
\label{sec:never}
To identify ``cultural redlines''--boundaries where AI representation is considered unwelcome-- respondents were asked: ``Are there aspects of your culture that you believe should NEVER be represented in generative AI tools?'' Across the 13 countries surveyed, more than 20\% of participants responded ``yes'' with percentages reaching 30\%+ in the majority of countries (see  Table \ref{tab:cultural_redlines}). 
This indicates a significant global consensus that certain cultural domains should remain outside the reach of GenAI representation. 

\begin{table}[h!]
    \centering
    \footnotesize
    \caption{Cultural Redlines in GenAI}
    \label{tab:cultural_redlines}
    \begin{tabularx}{\textwidth}{l c c c c X X}        \toprule
        \multicolumn{2}{l}{\textbf{}} & 
        \multicolumn{3}{c}{\textbf{\begin{tabular}[t]{@{}c@{}}Are There Social Identities \\ to Never Represent?\end{tabular}}} & 
        \multicolumn{2}{c}{\textbf{\begin{tabular}[t]{@{}c@{}}Specific Social Identities \\ to Never Represent\end{tabular}}} \\
        \cmidrule(lr){1-2} \cmidrule(lr){3-5} \cmidrule(lr){6-7}
        Country & N & Yes & No & Not sure & Example & English Translation \\
        \midrule
        Brazil & 464 & 28\% & 40\% & 32\% & ``[...] luta e a resistência dos povos indígenas e afro-brasileiros. [...] elementos como o samba ou o Carnaval de maneira comercial e descontextualizada [...].'' & \textit{ [...] the struggles and resistance of indigenous and Afro-Brazilian peoples. [...] elements such as samba or Carnival in a commercial and decontextualized manner [...]} \\
        \midrule
        Cameroon & 56 & 43\% & 34\% & 23\% & ``language, ancestral libations'' & \\
        \midrule
        France & 510 & 29\% & 37\% & 34\% & ``L’église la religion'', ``Des objets personnels'' & \textit{The church, religion}, \textit{Personal objects} \\
        \midrule
        Germany & 504 & 33\% & 37\% & 30\% & ``da durch KI immer wieder Fake-Bilder auftauchen sollten keine Menschen abgebildet werden, ganz egal in welcher Situation'' & \textit{Since fake images are constantly appearing due to AI, no people should be depicted, regardless of the situation.} \\
        \midrule
        India & 514 & 39\% & 39\% & 21\% & ``Stereotypical representations of my country'' &  \\
        \midrule
        Indonesia & 484 & 35\% & 43\% & 22\% & ``Baik, mungkin beberapa sejarah yang menurut saya tidak pernah atau lebih baik tidak ditampilkan, seperti pembantaian atau, lirik lagu yang berisi tentang secuil sejarah kelam'' & \textit{Okay, maybe some historical events that I think should never be shown, or are better left unshown, such as massacres or song lyrics that contain snippets of dark history.} \\
        \midrule
        Italy & 529 & 41\% & 32\% & 28\% & ``la propria identità'' & \textit{your own identity}\\
        \midrule
        Japan & 320 & 35\% & 30\% & 35\% & 特定の人物を使った関係のもの & \textit{Relationships involving specific people} \\
        \midrule
        Mexico & 302 & 23\% & 41\% & 36\% & ``Tradiciones y lengua'' & \textit{Traditions and language} \\
        \midrule
        Nigeria & 410 & 34\% & 45\% & 22\% & ``prayers'' & \\
        \midrule
        S. Korea & 505 & 28\% & 43\% & 29\% & ``인간성의 훼손 유의'' & \textit{Be mindful of the potential for damage to human dignity.} \\
        \midrule
        UAE & 533 & 42\% & 33\% & 25\% & \textarabic{الأهرامات والآثار المصريه} & \textit{Pyramids and Egyptian monuments} \\
        \midrule
        U.S. & 498 & 32\% & 35\% & 33\% & ``Spiritual and historical matters'' & \\
        \bottomrule
    \end{tabularx}
\end{table}

In open-ended follow-up questions, participants who answered ``yes'' were asked to give examples of what aspects of culture should never be represented. In Europe and Brazil, respondents frequently cited ``music'' and ``art,'' while also underscoring the importance of personal privacy. Respondents in Germany, South Korea, and the USA highlighted ``sensitive historical events'' and ``historical traumas'' in their responses, suggesting that GenAI models may lack the nuance required to handle complex collective memories. In some cases participants specifically named people or events that should NEVER be represented (e.g. ``slavery'' in the United States, or ''Hitler'' in Germany). An overarching theme across all countries was the sensitivity of cultural aspects related to \textit{Religion or Tradition}. This included religious artifacts, symbols, sacred practices, and language. Respondents expressed challenges with AI being insensitive and its capacity for misinterpretation (including sacred texts) through distorting identity. Participants equally pointed to concerns about human dignity and the potential for fake images, in one case stating that images of \textit{people} should not be depicted at all. Table \ref{tab:cultural_redlines} provides examples of these ``redlines'' in the original survey languages, with English translation.

\section{Discussion}

Certain aspects of culture emerged as salient for particular countries and certain trends remained consistent globally. Both can be leveraged to help understand and prioritize what aspects of culture are considered most important and sensitive.

\subsection{Culture and Practical Implications for GenAI Deployment}
These findings have implications for how we consider and prioritize culturally related interactions with GenAI. They can also help how we differentiate and evaluate culture within model responses in different regions. These findings underscore the need for GenAI developers to understand that expectations around representation are not uniform. The granular discussions about culture in the open-ended questions highlight the need to  not generalize assumptions and prioritize deep knowledge of cultural context. This need for deep understanding should influence fine-tuning of models, evaluation metrics and ultimately lead to better user experience. In terms of what users expressed as relevant to their identities, \textit{religion or tradition} was consistently either the top or of the top 3 attributes cited in each country. This demonstrates the importance of identities related to religious or traditional affiliation. Likewise, the artifacts cited in the open text responses for the question of \textit{what kinds of artifacts, practices, or other attributes do you consider most important?} - denote the importance of temples and other religious iconography. Furthermore in Section \ref{sec:never}, responses to the question about cultural redlines and what should NEVER be represented included religion and religious artifacts as being particularly sensitive. These questions confirm each other in how prominent the categories of religious affiliation are. Following \textit{religion or tradition}, social identities such as \textit{ethnicity}, \textit{nationality} and \textit{area where I live}, all related to geographic belonging, were of particular importance. We argue that developers should expand the view of what cultural studies in GenAI entails. The current body of work that focuses on understanding culture in the context of GenAI primarily focuses on culture from a geographic perspective, comparing the differences at the country or region level. These findings support including other characteristics as critical for how people understand their culture and what GenAI developers may need to pay attention to. 

\subsection{Navigating Nuance in Cultural Imaginaries}
We saw differences in what participants in different countries emphasized as being of importance in their own understanding of culture. Results from the survey question, \textit{what is your understanding of the word culture} - cited social practices and customs more frequently than other aspects. However, specific countries drew lines differently in what words they highlighted as the meaning of culture. European countries tended to include specific categories of artifacts, such as \textit{music}, \textit{art}, \textit{literature}, and \textit{history}. Countries in Asia included keywords such as \textit{country}, \textit{nation}, \textit{ancestors}, \textit{heritage}; while Sub-Saharan Africa centered on words like \textit{language} and \textit{practices}. All countries surveyed included words like \textit{customs}, \textit{traditions}, \textit{people}, that show the association of culture with people and social conventions. The differences between countries highlight how social conventions manifest in participants' imaginaries about culture. 

It is unclear whether these nuances are considered both by AI developers and whether they are present in the training data that influences what models are able to generate in terms of cultural representation \citep{qadri2025confusing, prabhakaran2022cultural, benkler2023assessing}. 
The concern is that this could potentially lead to harms that may be prioritized differently across different geographies. Looking at variations in what social identities participants in the different countries surveyed considered important and sensitive also confirms the nuance with respect to cultural imaginaries. Identities that are more closely linked with inherited characteristics tend to be considered both important and sensitive. However, social markers of status (e.g. socioeconomic status, education) or health are considered to be particularly sensitive in certain geographies. These subtle differences across countries are unlikely to be handled differently in the development process, where evaluations tend to be broader and only reflect the values of the country in which developers are situated \citep{vargaslarge}.

\section{Recommendations for Culture in GenAI development}

In recognition of the global reach of these models and related technologies, we recommend making culture front and center in the development process. Based on the findings discussed in this paper, we outline a methodological approach below that includes 4 pillars: awareness, participation, multi-facetedness, and nuance.

\subsection{Awareness: Include People's Perspectives}

Methods such as surveys, should be an integral practice in the development cycle. This method can offer at least an empirically-grounded baseline for how to approach certain salient cultural concepts during model training and post-training. Given that technology frequently becomes a part of culture, understanding and including global perspectives and imaginaries of technologies and their use are necessary for a complex nuanced user base. Our paper findings provide evidence for culture as being a multi-faceted feature of the human experience. In current model development practices this is an after-thought of model behavior, in that it is captured by virtue of the data on whiche the model was trained. We recommend re-situating culture and, specifically, the perspectives of people, as part of the development process. As such, building an awareness of people's perspectives means incorporating methods like surveys, interviews, or workshops as a necessary ``step'' in the process. 

\subsection{Participation: Include Users in Development Processes}
While the above pillar tackles approaches to building an awareness of the diverse perspectives of a user base, we also recommend that users participate more directly in development processes. A critical challenge identified is that the minority of researchers developing these models lack expertise in the cultures of the global majority. To bridge this gap, participatory approaches such as red teaming and data collection should be prioritized for a more thorough understanding in these models of the experience from communities centered around specific social identities. To illustrate this pillar, we outline three possible areas where users can participate: 

\begin{itemize}
    \item \textbf{Dimension-Based Judge Models:} Standard Reinforcement Learning from Human Feedback (RLHF) pipelines should be augmented with domain-specific raters that reflect a range of social and cultural identities. For example, given that participants expressed concerns about AI's capacity to misinterpret sacred texts, evaluation sets must be curated by community leaders from the specific groups identified as salient in our survey (e.g., consulting local religious scholars for the UAE context).
    \item \textbf{Community-Led Norms:} Rather than inferring values from internet data, which is known to be distorted, development cycles should integrate direct feedback loops from the communities who claimed ownership over these cultural aspects. Using the Religion/Tradition dimension as an example, this could mean reaching out to custodians of religious communities to ensure that the definitions of ``sacred" are defined from within the community. Community-led participatory design in AI is a rich area of research with precedent for this process \citep{martin2020participatory, young2024participation}.
    \item \textbf{Collaborative Audits:} Once models are released, collaborating with communities to audit outcomes should be a critical step in the process. This goes beyond defining norms through feedback loops, but rather soliciting community engagement in its own right. Algorithmic audits with community engagement have been proposed and are used in many contexts \citep{raji2020closing, lam2022end}. 
\end{itemize}

\subsection{Multi-facetedness: Culture is Not One-size-fits-all}
As noted in our findings, user sensitivities to cultural representation vary across global regions and demographics. Moreover in current practices, culture in AI is still steeped in AI rationality and understanding the world through a computational lens. Whether AI is being rolled out in the West or including the Global majority, the problem still remains that AI is limited by its own computational reasoning which consists of turning culture and language into discreet categories that are computationally reasoned over. As noted in \citep{qadri2025confusing} 
this is a fundamentally flawed understanding of how both language and culture operate. As such, we recommend a more multi-faceted approach to adapting models and AI workflows that can be tailored to specific dimensions of culture, such as region.   
\begin{itemize}
    \item \textbf{Dynamic Configuration:} We propose the use of region-specific ``cultural configuration files'' during inference. Instead of a global safety threshold, the model should utilize the sensitivity heatmaps generated by this study to dynamically up-weight or down-weight refusal probabilities based on the user's locale.
    \item \textbf{Contextual Adaptation:} In markets where specific identities, such as religion, are paramount, the ``temperature'' for creative variance on social identity related queries should be minimized to prevent hallucination, whereas in regions prioritizing artistic expression, the constraints may be adjusted to allow for respectful interpretation.
\end{itemize}

\subsection{Nuance: Prioritize Specific Cultural Dimensions}
Based on our survey findings, \textit{religion and tradition} emerged as a universally critical component of cultural identity, consistently ranking in the top three attributes across all regions surveyed. Furthermore, participants frequently cited religious artifacts, prayers, and sacred texts as specific aspects that should effectively be ``redlined" or handled with extreme care. A way to operationalize these findings is moving beyond generic safety filters towards \textbf{``Sensitivity" Frameworks} focused on cultural dimensions that emerge as particularly sensitive in the context of AI generated content. For example, a framework focused on a religious cultural dimension would prioritize the ontological protection of the divine and traditional over general conversational utility. To illustrate this framework, we outline a tiered approach.

Our data indicates that users distinguish between cultural aspects that are ``sensitive'' (requiring accuracy) and those that are ``prohibitive'' (requiring non-existence). A binary safety filter is insufficient; instead, we recommend adopting a tiered taxonomy related to sensitivity:

\begin{itemize}
    \item \textbf{Tier 1: The Prohibitive Redline.} 
    This tier includes sacred rituals, specific prayers, and depictions of the divine that users explicitly stated should \textit{never} be represented. For example, participants in Nigeria and the UAE specifically flagged ``prayers'' and ``monuments'' as areas of high sensitivity.
    \item \textbf{Tier 2: High-Fidelity Representation.} 
    This tier encompasses cultural religious artifacts where the harm lies in distortion or stereotyping rather than the act of generation itself. For these dimensions, the model must be constrained to generate content only when it can guarantee a high degree of historical and theological accuracy, potentially utilizing Retrieval-Augmented Generation (RAG) backed by authoritative sources.
\end{itemize}
Such labels can be utilized as part of the broader methodological approach discussed in these recommendations that includes using surveys, tailored judge models, community-led norms, and nuanced taxonomies to generate content that both respects and reflects a diverse user base.  

\section{Conclusion}
In this paper we presented a survey on perspectives and expectations of culture with respect to GenAI. The survey was rolled out globally in a phased approach and gathered feedback on participants' views on which elements of culture are most important to them as well as how these aspects are understood and surfaced by GenAI. The findings highlighted cultural differences across countries in both how participants' understood the word culture and what they found the most concerning with respect to cultural depictions in GenAI. Religion and Tradition were consistently in the top three aspects of culture, also supported by evidence from open ended questions. Across countries, demographics and level of familiarity with gen AI, participants stated that they viewed inaccurate results as more harmful than receiving no result. Likewise, across all countries, more than 20\% of participants, often more than 30\%, believed that there are aspects of culture that should never be represented by GenAI. Participants cited aspects such as religious artifacts, language, and their nationality as things that should never be represented. Alongside global trends, the survey findings also surfaced nuances between countries. Particularly in terms of what should never be represented, certain countries mentioned specific historical traumas and events that should be handled with care. 

\section{Limitations}
We highlight the inherent challenges of defining and measuring the complex, multifaceted concept of culture, as the survey's approach may not fully capture the dynamic and subconscious nature of cultural identity. The online survey methodology introduces potential sampling biases, as results may be skewed towards more digitally literate and engaged populations, despite quotas for age and gender. Further limitations arise from the survey's design; closed-ended questions can restrict responses to predefined categories that may not fully represent participants' views, as exemplified in Figure \ref{fig:country-aspects}. Additionally, there is a risk of social desirability bias, where respondents might offer more socially acceptable answers, as well as other biases common in surveys that require a high amount of self-awareness. Finally, despite translating the survey, linguistic nuances and varied interpretations of terms across cultures could impact the comparability and accuracy of the results, especially concerning sensitive topics.

\section{Ethical Considerations}
There are several ethical considerations to take into account. First, our work could be seen as suggesting that cultural differences \textit{should} be considered in GenAI responses. We can consider whether culture alongside other deeply human issues should be discussed at all. We also acknowledge that the terms used in the demographic questions are limited in how they capture the nuances of identities and may encode sociopolitical commitments that are also contested (Morning 2015). 
We must also discuss using locally dominant languages as part of the survey, which fails to provide material for considering expressions of culture beyond languages that are hegemonic in their respective countries. This may lend itself to a banalization of culture as discreet features that can be simply listed and collected. This runs the risk of stereotyping and appropriation of culture that disregards the rich histories from which such aspects emerge. 

We are aware that our methodology of aggregating data at a national level, while necessary for a global survey, can erase the diversity of minority groups within those nations and inadvertently create new, data-driven stereotypes. This approach also necessarily elides individual variation in cultural beliefs and practices. This bias can become even more pronounced since we, as researchers primarily from the Global North, collect cultural data from the global majority. Perhaps most critically, we must confront the dual-use potential of our findings. A detailed map of cultural sensitivities, in the wrong hands, could be weaponized to create targeted disinformation. By acknowledging these challenges, we aim to proceed with a greater sense of responsibility and contribute to a more equitable development of AI.

\section{Informed Consent}
Written informed consent was obtained from all participants between September 2024 and January 2025. Participants were informed about the aims of the study, their right to withdraw at any time without penalty, and consent was given for the use of their anonymised data for publication.

\bibliographystyle{unsrt}  
\bibliography{custom}

\end{document}